\documentclass[sigconf]{acmart}

\AtBeginDocument{%
  \providecommand\BibTeX{{%
    \normalfont B\kern-0.5em{\scshape i\kern-0.25em b}\kern-0.8em\TeX}}}

\copyrightyear{2023}
\acmYear{2023}
\setcopyright{rightsretained}
\acmConference[MM '23]{Proceedings of the 31st ACM International Conference on Multimedia}{October 29-November 3, 2023}{Ottawa, ON, Canada}
\acmBooktitle{Proceedings of the 31st ACM International Conference on Multimedia (MM '23), October 29-November 3, 2023, Ottawa, ON, Canada}
\acmDOI{10.1145/3581783.3613850}
\acmISBN{979-8-4007-0108-5/23/10}

\usepackage{multirow}
\usepackage{subfigure}
\usepackage{float}
\usepackage{balance}

\begin{document}

\title{Cross-modal Contrastive Learning for Multimodal Fake News Detection}


\author{Longzheng Wang}
\email{wanglongzheng@iie.ac.cn}
\affiliation{%
  \institution{Institute of Information Engineering, CAS}
  \institution{School of Cyber Security, UCAS}
  \country{China}
}

\author{Chuang Zhang$^*$}
\email{zhangchuang@iie.ac.cn}
\affiliation{%
  \institution{Institute of Information Engineering, CAS}
  \thanks{Corresponding author}
  \country{China}
}

\author{Hongbo Xu}
\email{hbxu@iie.ac.cn}
\affiliation{%
  \institution{Institute of Information Engineering, CAS}
  \country{China}
}

\author{Yongxiu Xu}
\email{xuyongxiu@iie.ac.cn}
\affiliation{%
  \institution{Institute of Information Engineering, CAS}
  \country{China}
}

\author{Xiaohan Xu}
\email{xuxiaohan@iie.ac.cn}
\affiliation{%
  \institution{Institute of Information Engineering, CAS}
  \institution{School of Cyber Security, UCAS}
  \country{China}
}

\author{Siqi Wang}
\email{wangsiqi2022@iie.ac.cn}
\affiliation{%
  \institution{Institute of Information Engineering, CAS}
  \institution{School of Cyber Security, UCAS}
  \country{China}
}

\renewcommand{\shortauthors}{Longzheng Wang et al.}

\begin{abstract}
Automatic detection of multimodal fake news has gained a widespread attention recently. Many existing approaches seek to fuse unimodal features to produce multimodal news representations. However, the potential of powerful cross-modal contrastive learning methods for fake news detection has not been well exploited. Besides, how to aggregate features from different modalities to boost the performance of the decision-making process is still an open question. To address that, we propose COOLANT, a cross-modal contrastive learning framework for multimodal fake news detection, aiming to achieve more accurate image-text alignment. To further capture the fine-grained alignment between vision and language, we leverage an auxiliary task to soften the loss term of negative samples during the contrast process. A cross-modal fusion module is developed to learn the cross-modality correlations. An attention mechanism with an attention guidance module is implemented to help effectively and interpretably aggregate the aligned unimodal representations and the cross-modality correlations. Finally, we evaluate the COOLANT and conduct a comparative study on two widely used datasets, Twitter and Weibo. The experimental results demonstrate that our COOLANT outperforms previous approaches by a large margin and achieves new state-of-the-art results on the two datasets.\footnote{Code and data are available at \url{https://github.com/wishever/COOLANT}}
\end{abstract}



\maketitle

\section{Introduction}
With the proliferation of Online Social Networks (OSNs) such as Twitter and Weibo, individuals can freely share daily information and express their opinions and emotions. However, the misuse of OSNs and the lack of proper supervision to verify the credibility of online posts have given rise to the widespread dissemination of considerable fake news~\cite{zubiaga2018detection}. Therefore, fake news detection has gained a widespread attention and has become a top priority recently.

\begin{figure}[t]
  \centering
  \includegraphics[scale=0.73]{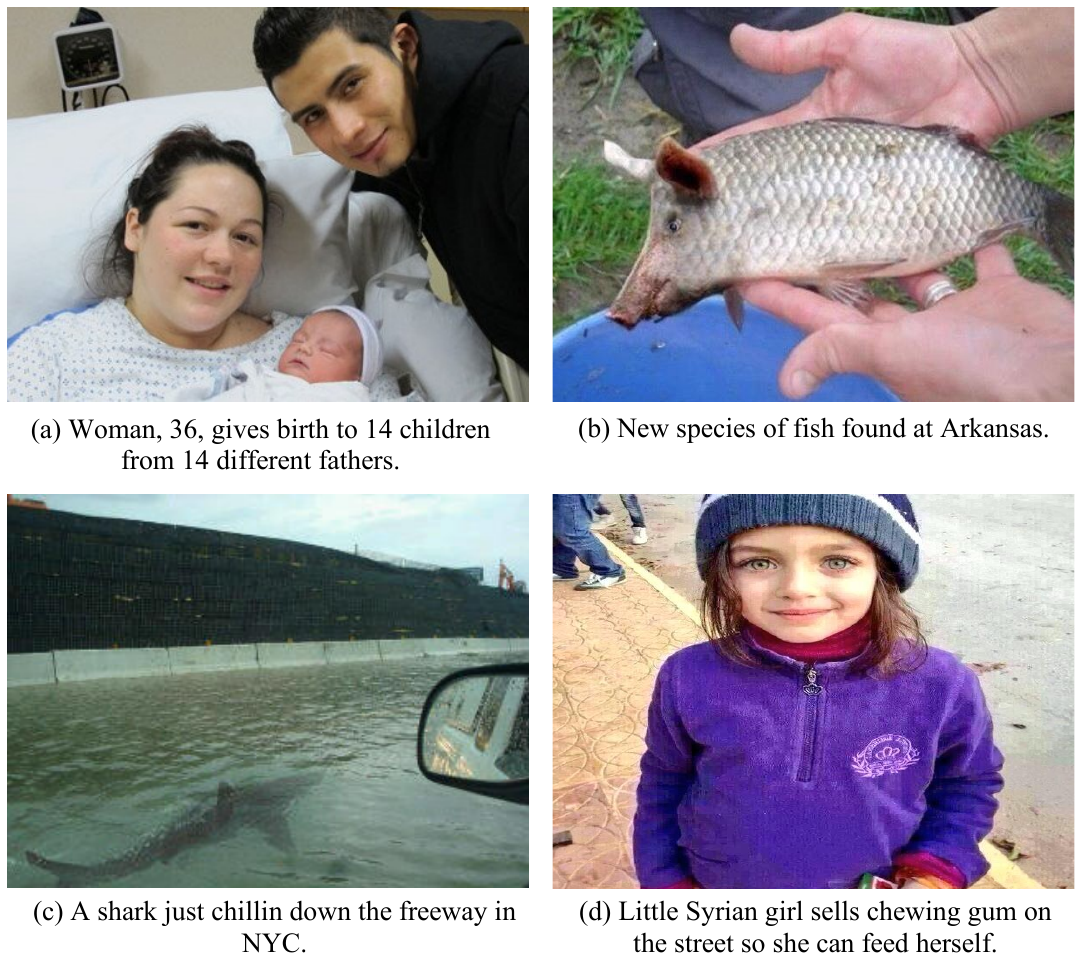}
  \caption{Some fake news from Twitter. \textbf{(a)} The image does not add substantial information while the textual contents indicate that it is possibly fake. \textbf{(b)} The doctored image suggests that it is probably fake news. \textbf{(c)} Both text and image demonstrate it is likely to be fake. \textbf{(d)} An instance of False Connection, which narrates a war-like situation but includes a happy emotion depicted in the image.
    }\label{fig:eg}
\end{figure}

Existing studies on automatic fake news detection mainly focus on textual content, either with traditional learning methods such as decision tree classifiers~\cite{liu2015real} or deep learning approaches such as convolutional neural networks (CNN)~\cite{yu2017convolutional}. However, most posts on social media commonly contain rich multimodal information, and the detection based on unimodal features is far from sufficient. Figure~\ref{fig:eg} shows some examples from Twitter illustrating the reasons why these four news items were determined to be fake. Recent works seek to fuse textual and visual features to produce multimodal post representations and then boost the performance of fake news detection~\cite{khattar2019mvae,wu2021multimodal}. Nevertheless, we argue that more advanced multimodal representation learning paradigms should be appropriately applied, since acquiring more sophisticated aligned unimodal representations and cross-modal features is a prerequisite for effective multimodal fake news detection. Besides, cross-modal features might not necessarily play a critical role in some cases~\cite{chen2022cross,singhal2022leveraging}. For instance, the textual contents in Figure~\ref{fig:eg}(a) are preposterous enough to indicate that it is fake. In contrast, the cross-modal information gap in Figure~\ref{fig:eg}(d) can help improve classification accuracy. Therefore, how features from different modalities affect the decision-making process and how we can make it more effective and interpretable remain open questions.


Recently, several contrastive learning-based multimodal pre-training methods have achieved great success, suggesting that contrastive learning may be a powerful paradigm for multimodal representation learning~\cite{he2020momentum,radford2021learning,jia2021scaling,li2021align,li2022blip,bao2022vlmo,yu2022coca}. A contrastive loss aims to align the image features and the text features by pushing the embeddings of positive image-text pair together while pushing those of negative image-text pair apart. It has been shown to be an effective objective for improving the unimodal encoders to better understand the semantic meaning of images and texts. While effective, the one-hot labels in contrastive learning penalize all negative predictions regardless of their correctness~\cite{li2021align,gao2022pyramidclip}. Therefore, this contrastive framework for multimodal fake news detection suffers from several key limitations: (1) A huge number of image-text pairs in fake news are inherently not matched (e.g. Figure~\ref{fig:eg}d), and the contrastive objective may overfit to those data and degrade the model's generalization performance; (2) Different image-text pairs may have potential correlation (especially in the case of different multimodal news about the same event), existing contrastive objectives directly treat those pairs as negative, which may confuse the model. Therefore, although these advanced technologies can be beneficial in multimodal representation learning, their application in multimodal fake news detection remains to be explored.



Taking the consideration above, we propose COOLANT, a \textbf{C}ross-m\textbf{o}dal C\textbf{o}ntrastive \textbf{L}e\textbf{a}rning framework for Multimodal Fake \textbf{N}ews De\textbf{t}ection. We utilize a simple dual-encoder framework to construct a visual semantics level and a linguistic semantics level. Then we use the image-text contrastive (ITC) learning objective to ensure the alignment between image and text modalities. As mentioned above, the contrastive learning framework utilized for detecting multimodal fake news is subject to certain constraints, primarily stemming from the one-hot labeling method. To alleviate this problem and further improve the alignment precision, we leverage an auxiliary task, called cross-modal consistency learning, to introduce more supervisions and bring in more fine-grained semantic information. Specifically, the contrastive learning objective ensures that the image-text pairs are in perfect one-to-one correspondence, and the consistency learning task can derive the potential semantic similarity features to soften the loss of negative samples (unpaired samples). After that, we feed the aligned unimodal representations into a cross-modal fusion module to learn the cross-modality correlations. Finally, we design an attention mechanism module to help effectively aggregate the aligned unimodal representations and the cross-modality correlations. Inspired by~\cite{chen2022cross}, we introduce an attention guidance module to quantify the ambiguity between text and image by estimating the divergence of their representation distributions, which can help guide the attention mechanism to assign reasonable weights to modalities. In this way, COOLANT can acquire more sophisticated aligned unimodal representations and cross-modal features, and then effectively aggregate these features to boost the performance of multimodal fake news detection.

The main contributions of this paper are as follows:
\begin{itemize}
    \item We propose COOLANT, a cross-modal contrastive learning framework for multimodal fake news detection, aiming to achieve more accurate image-text alignment.
    \item We soften the loss term of negative samples during the contrast process to ease the strict constraint, so as to make it more compatible with our task.
    \item We introduce an attention mechanism with an attention guidance module to help effectively and interpretably aggregate features from different modalities.
    \item We conduct experiments on two widely used datasets, Twitter and Weibo. Experimental results demonstrate that our model outperforms previous systems by a large margin and achieves new state-of-the-art results on the two datasets.
\end{itemize}

\section{Related Works}
\subsection{Fake News Detection}
\subsubsection{Unimodal Methods}
Existing unimodal fake news detection methods mainly rely on text content analysis or image content in posts. In text content analysis,~\cite{qian2018neural} propose a generative model to extract new patterns and assist fake news detection by analyzing past meaningful responses of users. TM~\cite{bhattarai2021explainable} exploits lexical and semantic properties of the text to detect fake news. Besides, verifying logical soundness~\cite{guo2018rumor}, capturing writing styles~\cite{potthast2017stylometric} or extracting rhetorical structure~\cite{conroy2015automatic} are also widely utilized to combat fake news. For image content,~\cite{jin2016novel} claim that there are noticeable discriminating features in the dissemination pattern of image content between real news and fake news. MVNN~\cite{qi2019exploiting} jointly leverages visual features in the spatial domain and image features in the frequency domain features for forensics. However, these approaches ignore cross-modal characteristics such as correlation and consistency, which may undermine their overall performance on multimodal news.

\subsubsection{Multimodal Methods}
More recently, several methods based on cross-modal discriminative patterns have been proposed to obtain superior performance in fake news detection. To learn the cross-modal characteristics, EANN~\cite{wang2018eann} leverages an additional event discriminator to aid feature extraction. MVAE~\cite{khattar2019mvae} introduces a multimodal variable autoencoder to learn probabilistic latent variable models and then reconstructs the original texts and low-level image features. MCAN~\cite{wu2021multimodal} stacks multiple co-attention layers to better fuse textual and visual features for fake news detection. However, studies in multimodal fake news detection have rarely considered the application of the recently emerged multimodal representation learning paradigms. Besides, some methods work on the principles of weak and strong modality. CAFE~\cite{chen2022cross} measures cross-modal ambiguity by evaluating the Kullback-Leibler (KL) divergence between the distributions of unimodal features. The learned ambiguity score then linearly adjusts the weight of unimodal and multimodal features before final classification. LIIMR~\cite{singhal2022leveraging} identifies the modality that presents more substantial confidence towards fake news detection. In this paper, we effectively leverage features from different modalities and make the decision process more interpretable.

\subsection{Contrastive Learning}
Recently, contrastive learning has achieved a great success in computer vision (CV)~\cite{he2020momentum,chen2020simple,chen2021empirical} and natural language processing (NLP)~\cite{gao2021simcse,yan2021consert}. It has also been adapted to vision-language representation learning. 
WenLan~\cite{huo2021wenlan} proposes a two-tower Chinese multimodal pre-training model and adapts MoCo~\cite{he2020momentum} into the cross-modal scenario. CLIP~\cite{radford2021learning} and ALIGN~\cite{jia2021scaling} demonstrate that dual-encoder models pretrained with contrastive objectives on massive noisy web data can learn strong image and text representations, which enable zero-shot transfer of the model to various downstream tasks. ALBEF~\cite{li2021align} employs a contrastive loss to effectively align the vision and language representations, followed by a cross-attention model for fusion. Furthermore, ALBEF~\cite{li2021align} presents a hard negative mining strategy founded on the contrastive similarity distribution, a method similarly employed by BLIP~\cite{li2022blip} and VLMo~\cite{bao2022vlmo}. CoCa~\cite{yu2022coca} conbines contrastive loss and captioning (generative) loss in an modified encoder-decoder architecture, which is widely applicable to many types of downstream tasks, and obtains a series of state-of-the-art performance. In this paper, we propose a cross-modal contrastive learning framework for multimodal fake news detection. In particular, our study utilizes an image-text contrastive (ITC) learning objective to effectively align the visual and language representations through a straightforward dual-encoder framework, thereby producing a unified latent embedding space. Moreover, we leverage an auxiliary cross-modal consistency learning task to measure the semantic similarity between images and texts, and then provide soft targets for the contrastive learning module.


\begin{figure*}[t]
  \centering
  \includegraphics[scale=0.49]{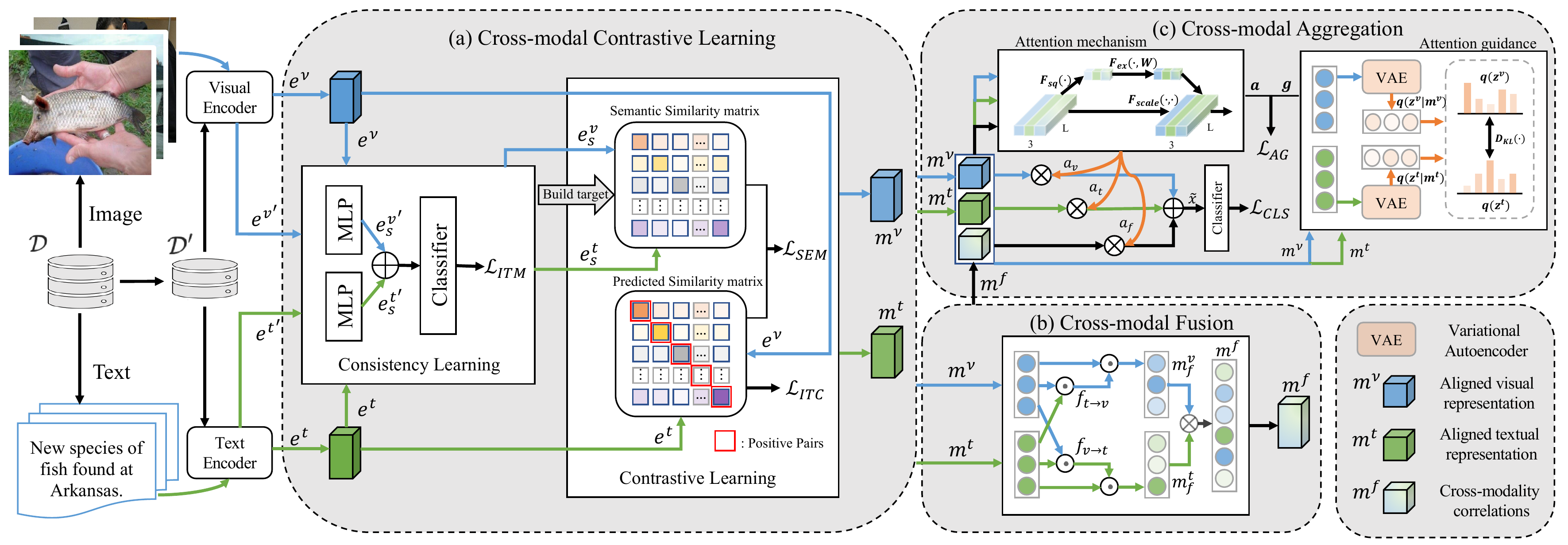}
  \caption{Model Architecture Overview of COOLANT. The model consists of three main modules: \textbf{(a) Cross-modal Contrastive Learning}: Given image-text pairs from the original dataset $\mathcal{D}$, we first extract unimodal features by the modal-specific encoder. Then we use the image-text contrastive learning objective to ensure the alignment between image and text modalities. To further improve the alignment precision, we leverage an auxiliary cross-modal consistency learning task based on the dataset $\mathcal{D}^{\prime}$ to provide semantic similarity matrixes as soft targets for the contrastive learning task. \textbf{(b) Cross-modal Fusion}: We feed the aligned unimodal representations into the cross-modal fusion neural network to learn the cross-modality correlations. \textbf{(c) Cross-modal Aggregation}: We apply an attention mechanism module to reweight the aligned unimodal representations and the cross-modality correlations. A VAE-based model is proposed to learn the ambiguity of different modalities and guide the assignment of the attention mechanism module.
}\label{fg:framework}
\end{figure*}

\section{Methodology}~\label{method}
In this section, we present our proposed framework COOLANT, that leverages the cross-modal contrastive learning task to align the features from image and text modalities. The overall model structure is illustrated in Figure~\ref{fg:framework}. Given image-text pairs, we first extract unimodal features by the modal-specific encoder ($\S$\ref{sec:encoder}). Then our method consists of three main components: the cross-modal contrastive learning module ($\S$\ref{sec:contrative}) for the alignment between image and text modalities, the cross-modal fusion module ($\S$\ref{sec:fusion}) for learning the cross-modality corrections, and the cross-modal aggregation module ($\S$\ref{sec:aggregation}) with an attention mechanism and an attention guidance for assigning reasonable attention scores to each modality, which then boosts the performance of multimodal fake news detection.

\subsection{Modal-specific Encoder} \label{sec:encoder}
Let each input multimodal news $\mathbf{x} = [x^v, x^t] \in \mathcal{D}$, where $x^v$, $x^t$ and $\mathcal{D}$ mean image, text and dataset, respectively. Since the modal-specific encoders are not the focus of this work, we leverage pre-training techniques to encode the image $x^v$ and the text $x^t$ into unimodal embedding $e^v$ and $e^t$, respectively.
\subsubsection{Visual Encoder} 
Given a visual content $x^v$, we utilize the pre-trained model ResNet~\cite{he2016deep} trained over the ImageNet database to extract regional features. The final visual embedding $e^v$ is obtained by using a fully connected layer to transform the regional features captured by ResNet.
\subsubsection{Text Encoder}
To precisely capture both semantic and contextualised representations, we adopt BERT~\cite{devlin2018bert} as the core module of our textual language model. Specifically, given a text $x^t$ with a set of words, each word is tokenized by a pre-prepared vocabulary, then we utilize BERT to obtain the aggregate sequence representation as temporal textual features. The final textual embedding $e^t$ is obtained by transforming the temporal textual features through a fully connected layer.

\subsection{Cross-modal Contrastive Learning}\label{sec:contrative}
Features from different modalities may have huge semantic gaps, so we adopt a more advanced multimodal representation learning paradigm, cross-modal contrastive learning, to align the features from different modalities by transforming the unimodal embeddings into a shared space. Specifically, we utilize a simple dual-encoder framework, establishing distinct visual semantics and linguistic semantics levels to construct a cross-modal contrastive learning module. As mentioned above, the one-hot labeling method in contrastive learning imposes a penalty on all negative predictions irrespective of their accuracy. Therefore, we propose to leverage an auxiliary cross-modal consistency learning task, which can help measure the semantic similarity between images and texts. The consistency learning module can provide semantic similarity matrixes as soft targets for the contrastive learning module. 


\subsubsection{Consistency Learning}
The cross-modal consistency learning is a binary classification task, which predicts whether a pair of image and text is positive (matched) or negative (not matched) given their multimodal feature. Specifically, we begin with crafting a new dataset $\mathcal{D}^{\prime}=[\mathcal{D}_{\text {pos}}, \mathcal{D}_{\text {neg }}]$ on the basis of $\mathcal{D}$, where a text-image pair is labeled $y^{\prime}$ = 1 if the textual and visual embeddings are from the same piece of real news, otherwise $y^{\prime}$ = 0. We feed the modal-specific encoders with $\mathbf{x}^{\prime} = [x^{v^{\prime}}, x^{t^{\prime}}] \in \mathcal{D}^{\prime}$ to obtain unimodal embeddings $e^{v^{\prime}}$ and $e^{t^{\prime}}$. The unimodal embeddings are projected to a shared semantic space via a modality-specific multilayer perceptron (MLP) to learn shared embeddings $e_{s}^{v^{\prime}}$ and $e_{s}^{t^{\prime}}$. Then, the shared embeddings are fed to an average pooling layer, followed by a full-connected layer as a binary classifier. We use the cosine embedding loss with margin $d$ as supervision:
\begin{equation}\label{eq:reg}
    \mathcal{L}_{ITM}=\left\{\begin{array}{ll}
    1-\cos (e_{s}^{v^{\prime}}, e_{s}^{t^{\prime}}) & \text{if } y^{\prime}=1 \\
    \max (0, \cos (e_{s}^{v^{\prime}}, e_{s}^{t^{\prime}})-d) & \text{if } y^{\prime}=0
\end{array}\right.
\end{equation}
where $\cos(\cdot)$ denotes the normalized cosine similarity and the margin $d$ is set as 0.2 due to empirical studies. With the gradients from back-propagation, the cross-modal consistency learning task can automatically learn a shared semantic space between multimodal embeddings, which can help measure their semantic similarity. The task can be in parallel learned with the contrastive learning task.

\subsubsection{Contrastive Learning}
For a batch of $N$ image-text pairs $\mathbf{x} = \{(x_{i}^v, x_{i}^t)\}_{i=1}^{N}$, where $i$ indicates the $i_{th}$ pair, the normalized embedded vectors $\{e_{i}^{v},e_{i}^{t}\}_{i=1}^{N}$ of the same dimension are obtained by the modal-specific encoders. The image-text contrastive learning aims to predict which of the $N \times N$ possible image-text pairings across a batch actually occurred. There are $N^2 - N$ negative image-text pairs within a training batch. Our contrastive losses are designed to achieve the alignment between visual representation and textual representation. Specifically, for the $i_{th}$ pair, the predicted vision-to-text similarity $\boldsymbol{p}_{i}^{v \rightarrow t}=\{p_{i j}^{v \rightarrow t}\}_{j=1}^{N}$ and text-to-vision similarity $\boldsymbol{p}_{i}^{t \rightarrow v}=\{p_{i j}^{t \rightarrow v}\}_{j=1}^{N}$ can be calculated through:
\begin{equation}
\begin{aligned}
    p_{i j}^{v \rightarrow t} & =\frac{\exp (\operatorname{sim}(e_{i}^{v}, e_{j}^{t}) / \tau)}{\sum_{j=1}^{N} \exp (\operatorname{sim}(e_{i}^{v}, e_{j}^{t}) / \tau)} \\
    \quad p_{i j}^{t \rightarrow v} & =\frac{\exp (\operatorname{sim}(e_{i}^{t}, e_{j}^{v}) / \tau)}{\sum_{j=1}^{N} \exp (\operatorname{sim}(e_{i}^{t}, e_{j}^{v}) / \tau)}
\end{aligned}
\end{equation}
where $\tau$ is a learnable temperature parameter initialized with 0.07 and the function $\operatorname{sim}(\cdot)$ conducts dot product to measure the similarity scores. The corresponding one-hot label vectors of the ground-truth $\boldsymbol{y}_{i}^{v \rightarrow t}=\{y_{i j}^{v \rightarrow t}\}_{j=1}^{N}$ and $\boldsymbol{y}_{i}^{t \rightarrow v}=\{y_{i j}^{t \rightarrow v}\}_{j=1}^{N}$, with positive pair denoted by 1 and negatives by 0, are used as the targets to calculate cross-entropy:
\begin{equation}
    \mathcal{L}^{v \rightarrow t}=-\frac{1}{N} \sum_{i=1}^{N} \sum_{j=1}^{N} y_{i j}^{v \rightarrow t} \log p_{i j}^{v \rightarrow t}
\end{equation}
Likewise, we can compute $\mathcal{L}^{t \rightarrow v}$ and then reach to:
\begin{equation}
    \mathcal{L}_{ITC}=\frac{\mathcal{L}^{v \rightarrow t}+\mathcal{L}^{t \rightarrow v}}{2}
\end{equation}

However, as mentioned above, this kind of hard targets may not be entirely compatible with multimodal fake news detection. To further improve the alignment precision, we use the consistency learning module to build a more refined semantic level as soft targets to provide more accurate supervisions.

\subsubsection{Build Soft Target}
Building upon the previous unimodal embeddings $e^v$ and $e^t$, the consistency learning module can project them to shared embeddings $e_s^v$ and $e_s^t$. For a batch of $N$ image-text pairs, we propose to leverage shared embeddings $\{(e_{s}^{v})_i,(e_{s}^{t})_i\}_{i=1}^{N}$ to build the semantic similarity matrix as the soft targets. Take the semantic vision-to-text similarity as an example. For the $i_{th}$ pair, the semantic vision-to-text similarity $\boldsymbol{s}_{i}^{v \rightarrow t}=\{s_{i j}^{v \rightarrow t}\}_{j=1}^{N}$  can be calculated through: 
\begin{equation}
    s_{i j}^{v \rightarrow t} =\frac{\exp (\operatorname{sim}((e_{s}^{v})_i, (e_{s}^{t})_j) / \tau)}{\sum_{j=1}^{N} \exp (\operatorname{sim}((e_{s}^{v})_i, (e_{s}^{t})_j) / \tau)}
\end{equation}
where $\tau$ is the temperature initialized at 0.07. Likewise, we can compute the semantic text-to-vision similarity $\boldsymbol{s}_{i}^{t \rightarrow v}$. 

\subsubsection{Semantic Matching Loss}
The semantic similarity $\boldsymbol{s}_{i}^{v \rightarrow t}$ and $\boldsymbol{s}_{i}^{t \rightarrow v}$ are used as the soft targets to calculate semantic matching loss. The semantic matching loss is hence the cross entropy between the predicted similarity and soft targets as:
\begin{equation}
    \mathcal{L}_{SEM}=-\frac{1}{2N} \sum_{i=1}^{N} \sum_{j=1}^{N} (s_{i j}^{v \rightarrow t} \log p_{i j}^{v \rightarrow t} + s_{i j}^{t \rightarrow v} \log p_{i j}^{t \rightarrow v})
\end{equation}

The final learning objective of the cross-modality contrastive Learning module is defined as:
\begin{equation}\label{eq:cl}
  \mathcal{L}_{CL} = \mathcal{L}_{ITC} + \lambda \mathcal{L}_{SEM}
\end{equation}
where $ \lambda $ controls the contribution of the soft targets mechanism. We jointly train the cross-modality contrastive learning module to produce the semantically aligned unimodal representations $m^v$ and $m^t$ as the input of the cross-modal fusion module and the cross-modal aggregation module.

\subsection{Cross-modal Fusion}\label{sec:fusion}
In order to capture the semantic interactions between different modalities, we adopt the cross-modal fusion module to learn cross-modality correlations~\cite{chen2022cross}. Specifically, given the aligned unimodal representations $m^v$ and $m^t$, we first obtain the inter-modal attention weights by calculating the association between unimodal representations:
\begin{equation}
\begin{array}{l}
f_{t \rightarrow v } =\operatorname{softmax}\left(\left[m^{v}\right]\left[m^{t}\right]^{T} / \sqrt{dim}\right) \\
f_{v \rightarrow t }=\operatorname{softmax}\left(\left[m^{t}\right]\left[m^{v}\right]^{T} / \sqrt{dim}\right) 
\end{array}
\end{equation}
where $dim$ denotes the dimension size of the unimodal representation. Then, we update the original unimodal embedding vectors by the inter-modal attention weights to obtain the explicit correlation features:
\begin{equation}
\begin{aligned}
{m}_f^{v} & = f_{t \rightarrow v } \times m^{v} \\
{m}_f^{t} & = f_{v \rightarrow t }\times m^{t}
\end{aligned}
\end{equation}
Finally, we use an outer product between ${m}_f^{v}$ and ${m}_f^{t}$ to define their interaction matrix ${m}^{f}$:
\begin{equation}
    {m}^{f} = {m}_f^{v} \otimes {m}_f^{t}
\end{equation}
$\otimes$ denotes outer product. The final correlation matrix ${m}^{f}$ is flattened into a vector.

\subsection{Cross-modal Aggregation}\label{sec:aggregation}
The input of the aggregation module is obtained by adaptively concatenating two sets of embeddings: the aligned unimodal representations $m^v$ and $m^t$ from the cross-modal contrastive learning module and the cross-modality correlations ${m}^{f}$ from the cross-modal fusion module. 
\subsubsection{Attention Mechanism}
Since not all modalities play an equal role in the decision-making process~\cite{singhal2022leveraging}, we propose to apply an attention mechanism module to reweight these features before their aggregation. Inspired by the success of Squeeze-and-Excitation Network (SE-Net)~\cite{hu2018squeeze,zhou2022multimodal}, we adopt an attention module to model modality-wise relationships and then weight each feature adaptively. Specifically, given these three $L \times 1$ features $m^v$, $m^t$ and $m^f$, we first concatenate them into one $L \times 3$ feature, where $L$ represents the length of the feature. We adopt global average pooling $F_{sq}(\cdot)$ to squeeze global modality information into a $1 \times 3$ vector. Then, we opt to employ a simple gating mechanism $F_{ex}(\cdot,W)$ with a sigmoid activation to fully capture modality-wise dependencies. The final output of the attention mechanism module is obtained by rescaling $F_{scale(\cdot,\cdot)}$ the $L \times 3$ feature, which will be used to obtain the attention weights $\mathbf{a}=\{a_{v}, a_{t}, a_{f}\}$. More details can refer to~\cite{hu2018squeeze}.
\subsubsection{Attention Guidance}
However, this kind of decision-making process is still at a black-box level, in which the network designs cannot explain why such weights are assigned to each modality. To make this process more interpretable, we utilize the Variational Autoencoder (VAE)~\cite{khattar2019mvae} to model the latent variable and form the attention guidance module. Specifically, given the aligned unimodal features $m^v$ and $m^t$, the variational posterior can be denoted as: $q(z \mid m)=\mathcal{N}(z \mid \mu(m), \sigma(m))$, in which the mean $\mu$ and variance $\sigma$ can be obtained from the modal-specific encoder. Considering the distribution over the entire dataset:
\begin{equation}
\begin{aligned}
q(z^{v}) &=\frac{1}{N} \sum_{i=1}^{N} q(z_{i}^{v} \mid m_{i}^{v})\\
q(z^{t}) &=\frac{1}{N} \sum_{i=1}^{N} q(z_{i}^{t} \mid m_{i}^{t})
\end{aligned}
\end{equation}
~\cite{chen2022cross} suggest when unimodal features present strong ambiguity, the fake news detector should pay more attention to cross-modal features, and vice versa, which is formulated as the cross-modal ambiguity learning problem. Following the definition of cross-modal ambiguity, we measure the ambiguity of different modalities in data sample $\mathbf{x}_i$ by the averaged Kullback-Leibler (KL) divergence between the distributions of unimodal features:
\begin{equation}
    g_{i}^{v \rightarrow t}=\left(\frac{\mathrm{D}_{K L}\left(q\left(z_{i}^{v} \mid m_{i}^{v}\right) \| q\left(z_{i}^{t} \mid m_{i}^{t}\right)\right)}{\mathrm{D}_{K L}\left(q\left(z^{v}\right) \| q\left(z^{t}\right)\right)}\right)
\end{equation}
where $D_{KL}(\cdot \| \cdot)$ stands for the KL divergence. Likewise, we can compute $g_{i}^{t \rightarrow v}$ and then reach to:
\begin{equation}
    g_{i}=\operatorname{sigmoid}(\frac{1}{2}(g_{i}^{v \rightarrow t}+g_{i}^{t \rightarrow v}))
\end{equation}
Then we can obtain the cross-modal ambiguity scores $\mathbf{g} = \{[1-g_i, 1-g_i, g_i]\}_{i=1}^N$. We develop another loss function $\mathcal{L}_{AG}$, which calculates the logarithmic difference between the attention weights $\mathbf{a}=\{a_{v}, a_{t}, a_{f}\}$ from the attention mechanism module and the ambiguity scores $\mathbf{g}$:
\begin{equation}
    \mathcal{L}_{AG} = D_{KL}(\mathbf{a} \| \mathbf{g})
\end{equation}
By minimizing $\mathcal{L}_{AG}$, the attention mechanism module learns to assign reasonable attention scores to modalities which means that the module assigns each modality based on the ambiguity of different modalities.
\subsubsection{Classifier}
Given the unimodal representations, the cross-modality correlations and the attention weights, the final representation $\tilde{\mathbf{x}}$ can be calculated through:
\begin{equation}
    \tilde{\mathbf{x}}=(a_{v} \times m^{v}) \oplus(a_{t} \times m^{t}) \oplus(a_{f} \times m^{f})
\end{equation}
where $\oplus$ represents the concatenation operation. Then, we feed it into a fully-connected network to predict the label:
\begin{equation}
    \hat{y}=\operatorname{softmax}(MLP(\tilde{\mathbf{x}}))
\end{equation}
We use the cross-entropy loss function as:
\begin{equation}
    \mathcal{L}_{CLS}=-(y \log (\hat{y})+(1-y) \log (1-\hat{y}))
\end{equation}
where $y$ denotes the ground-truth label. The final learning objective of the cross-modality aggregation module is defined as:
\begin{equation}\label{eq:ca}
  \mathcal{L}_{CA} = \mathcal{L}_{CLS} + \gamma \mathcal{L}_{AG}
\end{equation}
where $ \gamma $ controls the the ratio of $\mathcal{L}_{AG}$. We jointly train the cross-modality aggregation module to assign reasonable attention scores for each modality, and effectively leverage information from all modalities to boost the performance of multimodal fake news detection.

The final loss function for COOLANT is defined as the combination of the consistency learning loss in Eq.~\ref{eq:reg}, the contrastive learning loss in Eq.~\ref{eq:cl} and the cross-modal aggregation learning loss in Eq.~\ref{eq:ca}:
\begin{equation}
    \mathcal{L} = \mathcal{L}_{ITM} + \mathcal{L}_{CL} + \mathcal{L}_{CA}
\end{equation}

\begin{table*}[t]
\caption{Performance comparison between COOLANT and other methods on Twitter and Weibo datasets. Our method achieves the highest Accuracy among these methods, and its Precision, Recall, and F1-score also exceed most of the compared methods.}
\centering
\setlength{\tabcolsep}{3mm}{
\begin{tabular}{ccccccccc}
\toprule
\multirow{2}{*}   & \multirow{2}{*}{Method} &\multirow{2}{*}{Accuracy} &\multicolumn{3}{c}{Fake News}  & \multicolumn{3}{c}{Real News}\\
   \cline{4-9} &  & & Precision   & Recall  & F1-score & Precision  & Recall  & F1-score   \\
\midrule
\multirow{9}{*}{Twitter} 
& EANN   & 0.648  & 0.810   & 0.498  & 0.617   & 0.584     & 0.759  & 0.660  \\
& MVAE  & 0.745  & 0.801   & 0.719  & 0.758   & 0.689     & 0.777  & 0.730  \\
& MKEMN  & 0.715  & 0.814   & 0.756  & 0.708   & 0.634     & 0.774  & 0.660  \\
& SAFE  & 0.762   & 0.831    & 0.724    &0.774   & 0.695   & 0.811   & 0.748  \\
& MCNN  & 0.784  & 0.778   & 0.781    & 0.779   & 0.790  & 0.787   & 0.788  \\
& MCAN   & 0.809  & \textbf{0.889}   & 0.765   & 0.822   & 0.732 & 0.871    & 0.795   \\
& CAFE  & 0.806 & 0.807 & 0.799 & 0.803 & 0.805 & 0.813    & 0.809  \\
&LIIMR    & 0.831    & 0.836   & 0.832   & 0.830  & 0.825    & 0.830   & 0.827  \\
& COOLANT & \textbf{0.900}   & 0.879 & \textbf{0.922}   & \textbf{0.900}   & \textbf{0.923} & \textbf{0.880}& \textbf{0.901}  \\ 
\midrule
\multirow{11}{*}{Weibo} 
& EANN  & 0.827 & 0.847 & 0.812  & 0.829   & 0.807   & 0.843  & 0.825 \\
& MVAE    & 0.824    & 0.854   & 0.769  & 0.809    & 0.802  & 0.875   & 0.837 \\
& MKEMN  & 0.814  & 0.823   & 0.799  & 0.812   & 0.723     & 0.819  & 0.798  \\
&SAFE     &0.816   &0.818  &0.815  &0.817  &0.816  &0.818  &0.817 \\
& MCNN  & 0.823  & 0.858   & 0.801    & 0.828   & 0.787  & 0.848   & 0.816  \\
 & MCAN      & 0.899    & 0.913   & 0.889   & 0.901   & 0.884   & 0.909   & 0.897   \\
 & CAFE    & 0.840     & 0.855  & 0.830  & 0.842  & 0.825   & 0.851  & 0.837    \\
&LIIMR    & 0.900    & 0.882   & 0.823   & 0.847    &  0.908  & 0.941  & \textbf{0.925}  \\
& FND-CLIP   & 0.907  & 0.914 & 0.901  & 0.908 & 0.914 & 0.901  & 0.907 \\
& CMC & 0.908 & \textbf{0.940} & 0.869  & 0.899 & 0.876 & \textbf{0.945} & 0.907 \\
& COOLANT & \textbf{0.923}   & 0.927 & \textbf{0.923}   & \textbf{0.925}  & \textbf{0.919} & 0.922 & 0.920 \\
\bottomrule
  \end{tabular}}
  \label{tb:comparison}
\end{table*}

\section{Experiments}
\subsection{Experimental Configurations}
\subsubsection{Datasets}
Our model is evaluated on two real-world datasets: Twitter~\cite{boididou2018detection} and Weibo~\cite{jin2017multimodal}. The Twitter dataset was released for Verifying Multimedia Use task at MediaEval. In experiments, we keep the same data split scheme as the benchmark~\cite{boididou2018detection,chen2022cross}. The training set contains 6, 840 real tweets and 5, 007 fake tweets, and the test set contains 1, 406 posts. The Weibo dataset collected by~\cite{jin2017multimodal} contains 3749 fake news and 3783 real news for training, 1000 fake news and 996 real news for testing. In experiments, we follow the same steps in the work~\cite{jin2017multimodal,wang2018eann} to remove the duplicated and low-quality images to ensure the quality of the entire dataset.

\subsubsection{Baseline}
We compared our proposed COOLANT model with the following strong baselines:
\begin{itemize}
    \item \textbf{EANN}~\cite{wang2018eann}, which is a GAN-based model that aims to remove the event-specific features.
    \item \textbf{MVAE}~\cite{khattar2019mvae}, which uses a variational autoencoder coupled with a binary classifier to learn shared representations of text and image.
    \item \textbf{MKEMN}~\cite{zhang2019multi}, which exploits the external knowledge-level connections to detect fake news.
    \item \textbf{SAFE}~\cite{zhou2020mathsf}, which measures cross-modal similarity for fake news detection.
    \item \textbf{MCNN}~\cite{xue2021detecting}, which incorporates textual features, visual tampering features and cross-modal similarity in fake news detection.
    \item \textbf{MCAN}~\cite{wu2021multimodal}, which stacks multiple co-attention layers to fuse the multimodal features.
    \item \textbf{CAFE}~\cite{chen2022cross}, which measures cross-modal ambiguity to help adaptively aggregate unimodal features and cross-modal correlations.
    \item \textbf{LIIMR}~\cite{singhal2022leveraging}, which leverages intra and inter modality relationships for fake news detection.
    \item \textbf{FND-CLIP}~\cite{zhou2022multimodal}, which uses two pre-trained CLIP encoders to extract the deep representations from the image and text.
    \item \textbf{CMC}~\cite{wei2022cross}, which transfers cross-modal correlation by a novel distillation method.
\end{itemize}

\subsubsection{Implementation Details}
The evaluation metrics include Accuracy, Precision, Recall, and F1-score. We use the batch size of 64 and train the model using Adam~\cite{kingma2014adam} with an initial learning rate of 0.001 for 50 epochs with early stopping. The $\lambda$ in the contrastive learning loss (Eq.~\ref{eq:cl}) and  the $\gamma$ in the cross-modal aggregation learning loss (Eq.~\ref{eq:ca}) are set to 0.2 and 0.5, respectively. All codes are implemented with PyTorch~\cite{paszke2019pytorch} and run on NVIDIA RTX TITAN.

\subsection{Overall Performance}
Table~\ref{tb:comparison} presents the performance comparison between COOLANT and other methods on Twitter and Weibo datasets. As shown in the table, COOLANT significantly outperforms all the compared methods on every dataset in terms of \textit{Acc} and \textit{F1-score}, which demonstrates the effectiveness of our proposed model. Specifically, COOLANT obtains a new state-of-the-art with an accuracy of \textbf{90.0\%} on Twitter dataset, achieving significant improvements with \textbf{6.9\%}. COOLANT also reaches an accuracy of \textbf{92.3\%}, achieving a new state-of-the-art on Weibo dataset, which is \textbf{1.5\%} higher than the previous best one.

Numerous approaches to fake news detection, such as EANN~\cite{wang2018eann} and MVAE~\cite{khattar2019mvae}, rely solely on the utilization of fused features obtained through either direct concatenation or attention mechanisms. Despite their widespread use, these fused features may lack the requisite discriminatory capability to effectively differentiate between real and fake news, mainly due to the fact that the separately extracted text and image features may not exist in the same semantic space. CAFE~\cite{chen2022cross} employs a cross-modal alignment approach for training encoder models capable of mapping textual and visual data into a shared semantic space. The fused features obtained from aligned text and image inputs are then utilized for classification purposes. However, the effectiveness of the encoder's encoding process may be hampered by a limited number of available datasets and the utilization of suboptimal labeling methods during training. This still results in a significant semantic gap between text and image features, which may impact overall classification performance. Our study differs from previous approaches by employing an image-text contrastive learning objective to achieve optimal alignment of visual and language representations. The findings of our study indicate that the proposed model is capable of acquiring highly sophisticated aligned unimodal representations, which is considered to be a crucial factor for successfully detecting multimodal fake news.

Note that FND-CLIP~\cite{zhou2022multimodal} and CMC~\cite{wei2022cross} are not evaluated on Twitter dataset. Since many tweets on Twitter dataset are related to a single event, which can easily lead to model overfitting. In contrast, our model can deal with this situation more effectively by the cross-modal contrastive learning module. In particular, the image-text contrastive learning task is beneficial in discerning between news items within the same event. Furthermore, the consistency learning task is able to extract event-invariant features to mitigate the effects of variations in the target, thereby improving the detection of fake news on newly emerged events. As a result, the incorporation of the cross-modal contrastive learning module in our approach has also contributed to its enhanced generalizability. This has ultimately led to its superiority over state-of-the-art methods, as demonstrated by its exceptional performance on the Twitter and Weibo datasets.


\begin{table}[!t]
\caption{Ablation study on the architecture design of COOLANT on two datasets.}
\centering
\setlength{\tabcolsep}{1mm}{
\begin{tabular}{ccccc}
\hline
\multirow{2}{*}  & \multirow{2}{*}{Method} &\multirow{2}{*}{Accuracy} &\multicolumn{2}{c}{F1 score} \\
  \cline{4-5}  & &  & Fake News & Real News \\
\hline
\multirow{6}{*}{Twitter}
&COOLANT & \textbf{0.900}  & \textbf{0.900} & \textbf{0.901}\\
&- w/o ITM  & 0.883  & 0.883   & 0.884   \\
&- w/o ITC  & 0.871  & 0.864   & 0.878   \\
&- w/o CMF  & 0.878  & 0.872   & 0.884   \\
&- w/o ATT  & 0.875  & 0.862   & 0.886   \\
&- w/o AGU  & 0.894  & 0.885   & 0.901   \\
\hline
\multirow{6}{*}{Weibo}
&COOLANT & \textbf{0.923}  & \textbf{0.925} & \textbf{0.920} \\
&- w/o ITM  & 0.912  & 0.912   & 0.911   \\
&- w/o ITC  & 0.896  & 0.895   & 0.896   \\
&- w/o CMF  & 0.909  & 0.912   & 0.907   \\
&- w/o ATT  & 0.904  & 0.903   & 0.904   \\
&- w/o AGU  & 0.906  & 0.907   & 0.906   \\
\hline
\end{tabular}}
\label{tb:ablation}
\end{table}

\subsection{Ablation Studies}
\subsubsection{Quantitative Analysis}
To evaluate the effectiveness of each component of the proposed COOLANT, we remove each one from the entire model for comparison. More specifically, the compared variants of COOLANT are implemented as follows: \textbf{1) w/o ITM:} we remove the consistency learning task and only use hard targets for the contrastive learning task to learn the aligned unimodal representations; \textbf{2) w/o ITC:} we remove the image-text contrastive learning task and use the consistency learning task to learn the aligned unimodal representations; \textbf{3) w/o CMF:} we remove the cross-modal fusion module and replace it with simply concatenating $m^v$ and $m^t$; \textbf{4) w/o ATT:} we remove the attention mechanism module and direct aggregate the three features to obtain final feature; \textbf{5) w/o AGU:} we remove the attention guidance module. 

Table~\ref{tb:ablation} shows the results of ablation studies. We can find that all variants perform worse than the original COOLANT, which demonstrates the effectiveness of each component. Besides, we have the following observations:
\begin{itemize}
    \item COOLANT w/o ITC yields the worst performance, indicating the necessity of acquiring more sophisticated aligned unimodal features for effective detection. Moreover, our study reveals that the image-text contrastive learning objective can facilitate optimal alignment of visual and language representations, which is crucial for enhancing the performance of the multimodal fake news detection task.
    \item The performance of COOLANT w/o ITM on Twitter dataset drops more noticeably than on Weibo dataset. As aforementioned, a considerable number of tweets in the Twitter dataset pertain to a single event, thereby impeding the efficacy of the contrastive learning framework due to the limitations of the one-hot labeling method. This result verifies that soft targets can help the model to maintain the event-invariant features and detect news related to the same event more effectively. Furthermore, the Weibo dataset has a larger scale than the Twitter dataset, implying that corpus scale can to some extent compensate for the noise in the dataset, as observed in ALIGN's prior findings~\cite{jia2021scaling}.
\end{itemize}


\subsubsection{Qualitative Analysis}
\begin{figure}[htbp]
  \centering
  \includegraphics[width=1\linewidth]{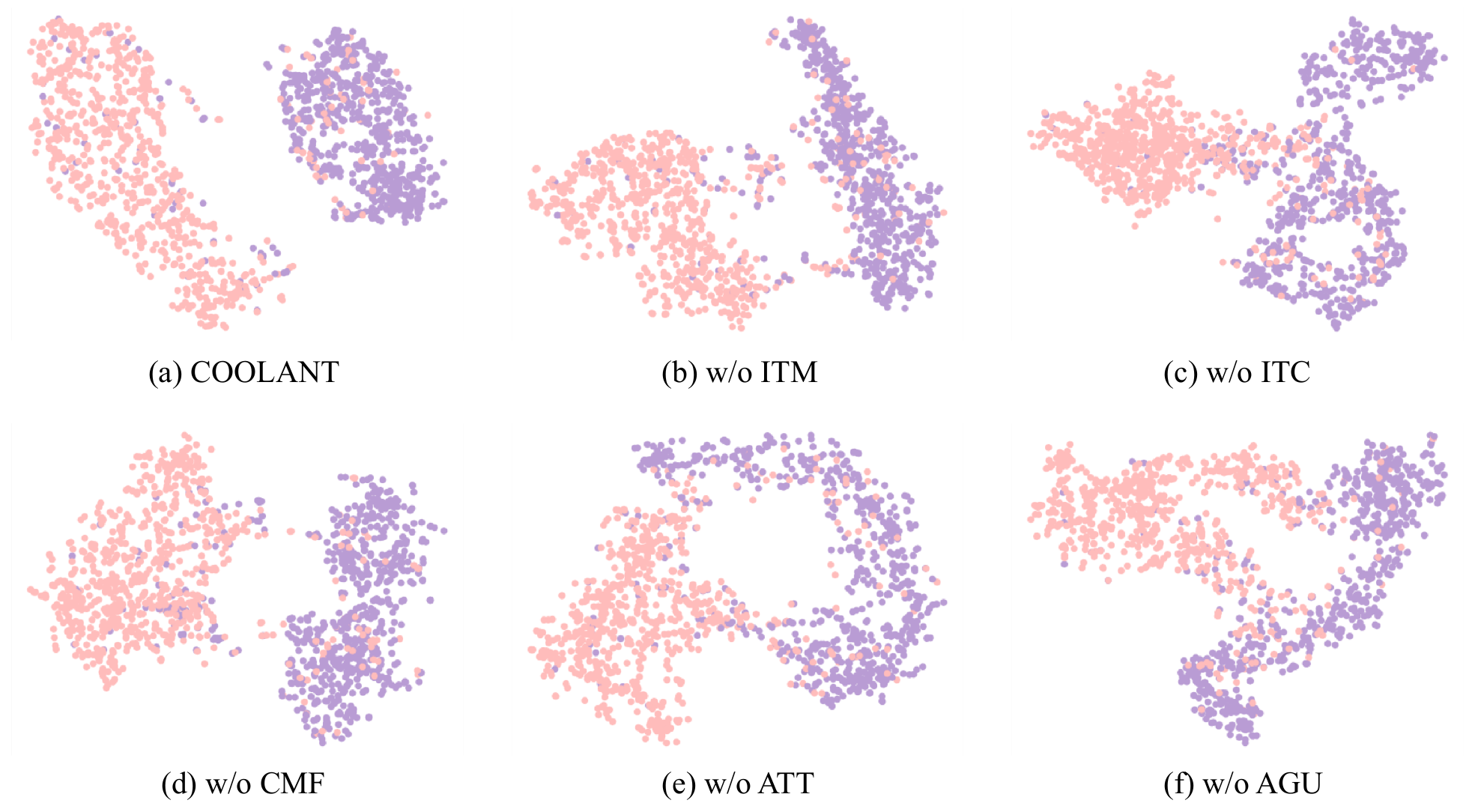}
  \caption{T-SNE visualizations of the features before classifier that are learned by COOLANT and its five variants on the test dataset of Weibo. Dots with the same color are within the same label.}
  \label{fg:tsne}
\end{figure}

Moreover, we further analyze the proposed method using t-SNE~\cite{van2008visualizing} visualizations of the features before classifier in Figure~\ref{fg:tsne}, which are learned by COOLANT and its five variants on the test dataset of Weibo.

From Figure~\ref{fg:tsne}, we can observe that the boundary of different label dots in COOLANT is more pronounced than that in its variants, revealing that the extracted features in COOLANT are more discriminative. Note that, as shown in Figure~\ref{fg:tsne}(c), many features learned by COOLANT w/o ITC are still easily misclassified, which indicates that the image-text contrastive learning task can obtain the characteristics of multiple modalities deeply and boost to distinguish fake news and real news. In addition, by comparing Figure~\ref{fg:tsne}(a), Figure~\ref{fg:tsne}(e) and Figure~\ref{fg:tsne}(f), we can see that effective and appropriate aggregation of features from different modalities can significantly improve the representation ability of the final features.

\section{Conclusion}
In this paper, we propose COOLANT, a novel cross-modal contrastive learning framework for multimodal fake news detection, which uses the image-text contrastive learning objective to achieve more accurate image-text alignment. To further improve the alignment precision, we leverage an auxiliary task to soften the loss term of negative samples during the contrast process. After that, we feed the aligned unimodal representations into a cross-modal fusion module to learn the cross-modality correlations. An attention mechanism with an attention guidance module is implemented to help effectively and interpretably aggregate features from different modalities. Experimental results on two datasets Twitter and Weibo demonstrate that COOLANT outperforms previous approaches by a large margin and achieves new state-of-the-art results on the two datasets.

\begin{acks}
The authors thank all the anonymous reviewers for their constructive comments. This work was supported by the National Key Research and Development of China (No. 2021YFB3100600), and Strategic Priority Research Program of Chinese Academy of Sciences (No. XDC02040400).
\end{acks}

\bibliographystyle{ACM-Reference-Format}
\balance
\bibliography{sample-base}


\end{document}